%% file: main.tex
\ifdefined\XeTeXversion\else\pdfoutput=1\fi
\documentclass{article}
\usepackage[T1]{fontenc}
\usepackage{times}
\usepackage{arxiv_preprint}

\usepackage{graphicx}
\usepackage{booktabs}
\usepackage{multirow}
\usepackage{amsmath,amssymb}
\usepackage{xcolor}
\usepackage{microtype}
\usepackage[shortlabels,inline]{enumitem}
\setlist[itemize]{noitemsep,leftmargin=*,topsep=0.3em}
\usepackage{xspace}
\usepackage[breaklinks]{hyperref}
\usepackage{url}
\hypersetup{hidelinks}
\usepackage[capitalize]{cleveref}
\crefname{section}{Sec.}{Secs.}
\crefname{appendix}{Appendix}{Appendices}
\Crefname{appendix}{Appendix}{Appendices}
\Crefname{section}{Section}{Sections}
\crefname{table}{Tab.}{Tabs.}
\Crefname{table}{Table}{Tables}
\crefname{figure}{Fig.}{Figs.}
\crefname{equation}{Eq.}{Eqs.}
\creflabelformat{equation}{#2(#1)#3}

\makeatletter
\DeclareRobustCommand\onedot{\futurelet\@let@token\@onedot}
\def\@onedot{\ifx\@let@token.\else.\null\fi\xspace}
\makeatother
\def\etal{\emph{et al}\onedot}

\newcommand{\ours}{TabDepth}
\newcommand{\direct}{\ours-D}
\newcommand{\prior}{\ours-P}
\newcommand{\best}[1]{\textbf{#1}}

\papertitle{Depth Completion as In-Context Tabular Regression}
\authorline{Chimata Anudeep\affmark{1}}
\affiliationline{\affmark{1}BITS Pilani, K.\,K. Birla Goa Campus\\[5pt]
Code: \url{https://github.com/einnewfeyhaw/tabdepth}}

\begin{document}
\maketitle
\input{sec/0_abstract}
\input{sec/1_intro}
\input{sec/2_related}

\input{sec/3_method}
\input{sec/4_experiments}
\input{sec/6_conclusion}

\bibliographystyle{plainnat}
\bibliography{main}

\clearpage
\renewcommand{\theHtable}{S\arabic{table}}
\renewcommand{\theHfigure}{S\arabic{figure}}
\section*{Appendix}
\addcontentsline{toc}{section}{Appendix}
\input{sec/X_suppl}

\end{document}

%% file: sec/0_abstract.tex
\begin{abstract}
Depth completion predicts a dense depth map from an image and a few depth measurements, and current zero-shot methods train dedicated networks for it on hundreds of thousands of images.
We ask whether this training is necessary, and cast depth completion as in-context tabular regression.
A tabular foundation model, trained only on synthetic tables, then predicts the depth of every pixel with no parameter updates.
Our method, TabDepth, needs no completion-specific training.
Using only pixel position, colour, and self-supervised image features, so that no component has seen depth, it reaches an RMSE of 0.117\,m on NYUv2 from 500 measurements, close to the zero-shot completion network OMNI-DC (0.111\,m) and well ahead of classical regressors given the same inputs (0.168\,m).
When we also give it the output of an off-the-shelf monocular depth network, TabDepth outperforms OMNI-DC on NYUv2, VOID, and all 20 synthetic-pattern settings on four unseen datasets, and it handles depth super-resolution and missing regions without any change.
An analysis of when the approach works shows that accuracy depends on how well the measurements cover the scene and whether they agree with it, which explains why OMNI-DC remains more accurate on real LiDAR and structure-from-motion inputs.
\end{abstract}

%% file: sec/1_intro.tex
\section{Introduction}
\label{sec:intro}

Depth completion predicts a dense depth map from an RGB image and sparse depth measurements, typically from LiDAR or structure-from-motion, and is a core component of autonomous driving, robotics, and 3D reconstruction.
Methods trained on a single dataset perform well in-domain but degrade under new sensors, measurement densities, or scene types~\cite{zuo2025omnidc}.
Recent zero-shot methods address this by training on large mixtures of datasets and sampling patterns, such as OMNI-DC~\cite{zuo2025omnidc} with 573K training images, or by building on large pretrained monocular depth and diffusion models~\cite{viola2025marigolddc,wang2025priorda}.
In all of them, the mapping from sparse measurements to dense depth is learned from depth data.

We ask whether such task-specific training is necessary.
Our starting point is tabular foundation models (TFMs)~\cite{muller2022pfn,hollmann2025tabpfnv2}, transformers pretrained exclusively on millions of synthetic tables.
Given a new table in which only some rows are labelled, a TFM predicts the labels of the remaining rows by in-context learning, without any parameter updates, and on small real-world tables it outperforms tuned gradient-boosted trees~\cite{hollmann2025tabpfnv2}.

\begin{figure*}[t]
\centering
\includegraphics[width=\linewidth]{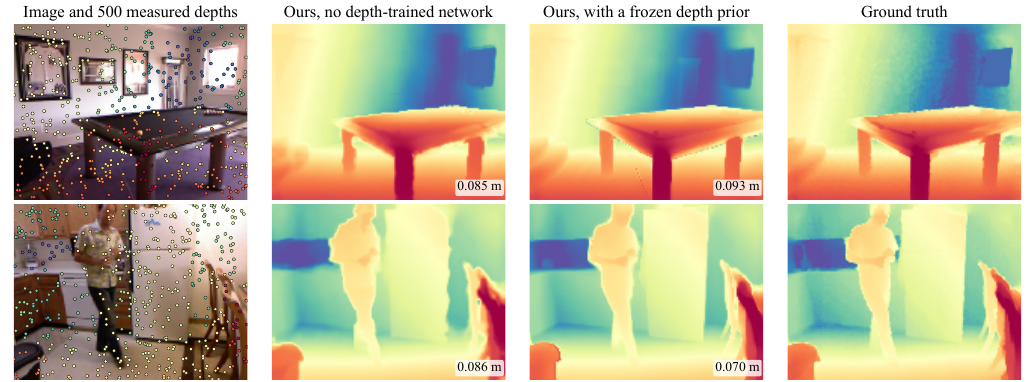}
\caption{\textbf{Depth from 500 measurements with a model trained only on synthetic tables.} Each pixel is a row of a table and the 500 measured depths (dots, left) are its labels; a frozen tabular foundation model predicts the depth of every other pixel. Second column: \direct{}, which uses pixel position, colour, and self-supervised image features, so no component has been trained on depth. Third column: \prior{}, which also uses a frozen monocular depth network. Numbers give the root-mean-square error (RMSE) of each map in metres; lower is better. NYUv2 test images.}
\label{fig:teaser}
\end{figure*}

We observe that depth completion can be written as exactly this problem.
Each pixel becomes a row, features from frozen image encoders become its columns, and the sparse depth measurements provide the labels.
A frozen TFM then predicts the depth of every remaining pixel.
Although the TFM has never seen an image or a depth map, it produces accurate dense depth (\cref{fig:teaser}).

We call this approach TabDepth and study two variants (\cref{sec:method}).
\direct{} (direct) uses only pixel position, colour, and self-supervised DINOv2 features~\cite{oquab2024dinov2}, so no component has been trained on depth.
\prior{} (prior-assisted) additionally uses a frozen monocular depth network, Depth Anything V2~\cite{yang2024dav2}, as a starting point, and the TFM predicts a correction to it.
Neither variant involves completion-specific training.

On NYUv2 with 500 measurements, \direct{} reaches an RMSE of 0.117\,m, close to OMNI-DC (0.111\,m) and ahead of G2-MonoDepth~\cite{wang2024g2} (0.122\,m).
\prior{} lowers the error to 0.108\,m and outperforms the published OMNI-DC on NYUv2, VOID, and all 20 synthetic-pattern settings on iBims-1, ARKitScenes, ETH3D, and DIODE (\cref{sec:benchmark}).
On VOID it also has lower error than the published Marigold-DC~\cite{viola2025marigolddc}, which guides a depth diffusion model by test-time optimisation.
The gain comes from the TFM itself: given identical inputs, tuned nearest-neighbour regression, gradient-boosted trees, and Gaussian processes have 32\% to 49\% higher error (\cref{sec:direct}).

Finally, we study what limits the accuracy of TabDepth.
TabDepth remains less accurate than OMNI-DC on real LiDAR (KITTI) and structure-from-motion points (ETH3D).
Controls with reference depth trace this gap to a small fraction of measurements that disagree with the true depth, especially when these errors are spatially clustered (\cref{sec:failures}).

\paragraph{Contributions.}
\begin{itemize}
    \item We formulate depth completion as in-context tabular regression, and show that a TFM trained only on synthetic tables predicts accurate depth without any depth-trained component; the best tuned classical regressor on identical inputs has 44\% higher error on NYUv2 at 500 points (\cref{sec:method,sec:direct}).
    \item We present TabDepth, a method with no completion-specific training that outperforms the published OMNI-DC on NYUv2, VOID, and all 20 synthetic-pattern settings across four datasets, and extends without modification to depth super-resolution and missing regions (\cref{sec:benchmark,sec:layouts}).
    \item We analyse how accuracy depends on the distance to the measurements, the depth network, and the choice of context, and identify disagreement between the measurements and the true depth as the main cause of the remaining failures (\cref{sec:analysis,sec:failures}).
\end{itemize}
Code and configurations are available at \url{https://github.com/einnewfeyhaw/tabdepth}.

%% file: sec/2_related.tex
\section{Related Work}
\label{sec:related}

\paragraph{Depth completion.}
Networks trained on one dataset, such as NLSPN~\cite{park2020nlspn}, CompletionFormer~\cite{zhang2023cformer}, and BP-Net~\cite{tang2024bpnet}, are accurate on that dataset and degrade when the sensor, the density, or the scene type changes~\cite{zuo2025omnidc}.
Zero-shot completion networks address the degradation by training on large mixtures of mostly synthetic datasets and sparsity patterns: G2-MonoDepth~\cite{wang2024g2}, DepthPrompting~\cite{park2024depthprompting}, OGNI-DC~\cite{zuo2024ognidc}, and OMNI-DC~\cite{zuo2025omnidc}.
Prior Depth Anything~\cite{wang2025priorda} trains a conditioned monocular model on top of a coarse alignment.
All of these need a completion training set.
A second group removes that requirement by optimising at test time: Marigold-DC~\cite{viola2025marigolddc} guides a depth diffusion model~\cite{ke2024marigold} with the sparse points, Lee~\etal~\cite{lee2025zeroshotdc} align an affine-invariant depth prior to them by optimisation, and recent work adapts a frozen 3D foundation model to each input through a small number of parameters~\cite{capa2026}, or makes test-time optimisation efficient enough for video~\cite{efftto2026}.
Our method keeps all network parameters fixed: the measurements fit a per-image scale and shift in the prior-assisted construction and provide the context of the tabular model.

\paragraph{Monocular depth as a prior.}
Relative depth networks such as MiDaS~\cite{ranftl2022midas} and Depth Anything V2~\cite{yang2024dav2} predict depth up to a scale and shift, and fitting these to sparse measurements is a standard completion baseline~\cite{zuo2025omnidc}.
We keep the network frozen and treat the error that the fit leaves as a regression problem over pixels.

\paragraph{Tabular foundation models.}
Prior-data fitted networks~\cite{muller2022pfn} are transformers trained on synthetic datasets drawn from a prior, so that a forward pass over a labelled context approximates the posterior predictive distribution of a query.
TabPFN~\cite{hollmann2023tabpfn,hollmann2025tabpfnv2} applies the idea to tables; TabICL~\cite{qu2025tabicl,tabiclv2} and Kumo-Tabular~\cite{kumotabular} extend it to larger tables.
TFMs have been applied outside tables, to time series~\cite{hoo2025tabpfnts}, signals and image denoising~\cite{tabpfnsignals}, and image classification from raw pixels~\cite{tfmnontabular}.
These applications classify or denoise; we apply a TFM to dense geometric prediction and compare it with trained vision models.

%% file: sec/3_method.tex
\section{Depth Completion as Tabular Regression}
\label{sec:method}

\subsection{Why a Tabular Model Can Predict Depth}
\label{sec:why}

We view depth completion as Bayesian inference.
Let $\mathcal{D}=\{(x_i,y_i)\}_{i=1}^{N}$ be the measurements, where $x_i$ describes a pixel and $y_i$ is its depth.
If the depths of an image are generated by an unknown function $\phi$ drawn from a prior $p(\phi)$, the optimal prediction at a query pixel $x$ is the posterior predictive distribution (PPD)
\begin{equation}
    p(y\mid x,\mathcal{D})=\int p(y\mid x,\phi)\,p(\phi\mid\mathcal{D})\,d\phi .
    \label{eq:ppd}
\end{equation}
For realistic priors this integral is intractable, which is one reason depth completion is usually learned from data.

Prior-data fitted networks (PFNs)~\cite{muller2022pfn} approximate \cref{eq:ppd} directly.
A transformer $q_\theta$ receives a labelled dataset $\mathcal{D}$ and a query $x$, and is trained on datasets sampled from the prior,
\begin{equation}
    \theta^\star=\arg\min_\theta\;\mathbb{E}_{\phi\sim p(\phi),\;\mathcal{D}\cup\{(x,y)\}\sim p(\cdot\mid\phi)}\big[-\log q_\theta(y\mid x,\mathcal{D})\big].
    \label{eq:pfn}
\end{equation}
M\"uller~\etal~\cite{muller2022pfn} show that this loss equals, up to a constant, the expected KL divergence between the PPD and $q_\theta$, so a network with enough capacity that minimises it recovers the PPD of its prior.
At test time, inference is a forward pass over the labelled context, and no parameter is updated.

Tabular foundation models are PFNs whose prior is a broad family of functions over tables, generated by random structural causal models~\cite{hollmann2025tabpfnv2,kumotabular}.
This raises the question that motivates our work.
Within one image, depth is a piecewise smooth function of where a pixel is and what it looks like: it varies slowly across a surface and jumps at object boundaries, which usually coincide with changes in appearance.
If such functions are well represented under the tabular prior, a TFM should approximate the PPD of depth given the measurements without ever having seen depth.
The rest of this section turns an image into a table so that this hypothesis can be tested.

\subsection{An Image as a Table}
\label{sec:formulation}

We are given an RGB image $I$ with pixel set $\Omega$ and measurements $\mathcal{S}=\{(p_i,d_i)\}_{i=1}^{N}$, where $p_i\in\Omega$ is a pixel and $d_i>0$ its measured depth.
We predict $\hat D(p)$ for all pixels.

\begin{figure*}[t]
\centering
\includegraphics[width=\linewidth]{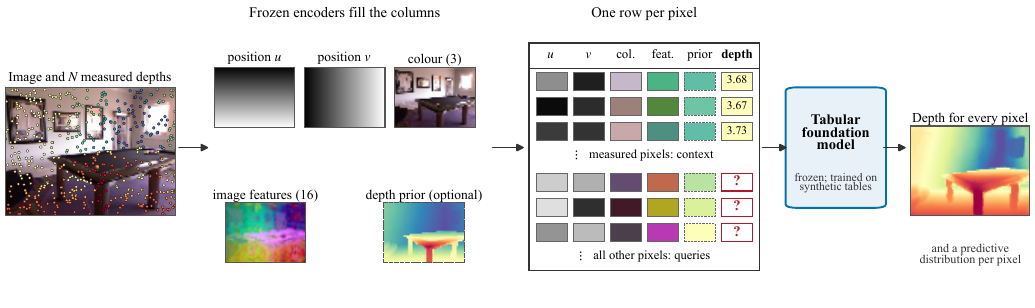}
\caption{\textbf{The method, on a real example.} Frozen encoders turn the image into per-pixel columns: position, colour, self-supervised image features, and optionally the output of a monocular depth network. Each pixel is one row. The $N$ measured pixels carry a depth label and form the context; every other pixel is a query. A frozen tabular foundation model predicts the label of each query row from the context, and nothing is trained.}
\label{fig:method}
\end{figure*}

We write the image as a table with one row per pixel.
The columns $z(p)\in\mathbb{R}^{F}$ are computed from the image by frozen encoders.
A row has a label only where a measurement exists.
The $N$ measured rows form the context $\mathcal{C}=\{(z(p_i),y_i)\}_{i=1}^{N}$, and every other pixel is a query.
A tabular foundation model (TFM) $T_\theta$ maps a context and a query row to a distribution over the label of the query:
\begin{equation}
    q\big(y \mid z(p),\,\mathcal{C}\big) \;=\; T_\theta\big(z(p);\,\mathcal{C}\big).
    \label{eq:tfm}
\end{equation}
The parameters $\theta$ come from pretraining on synthetic tables (\cref{eq:pfn}) and stay fixed, so \cref{eq:tfm} approximates the PPD of \cref{eq:ppd} for our table and is the entire prediction step (\cref{fig:method}).
The two constructions below differ in the columns $z$ and in the definition of the label $y$.

\subsection{Direct Construction}
\label{sec:directmethod}

The direct construction (\direct) uses no network trained on depth.
Its 21 columns are the pixel position $(u,v)$ (column and row coordinates scaled to $[0,1]$, with no positional encoding), the colour, and the first 16 principal components of the last-layer patch tokens of DINOv2~\cite{oquab2024dinov2}, computed per image and upsampled bilinearly.
The label is the log depth relative to the median measurement $\bar d$, $y_i=\log d_i-\log\bar d$, and the prediction is $\hat D(p)=\bar d\,\exp(\hat y(p))$.

\subsection{Prior-Assisted Construction}
\label{sec:residual}

A frozen monocular network~\cite{yang2024dav2} predicts a relative inverse depth map $r(p)$ from the image alone.
It captures the shape of the scene but not its metric scale: $r$ relates to the true inverse depth through an unknown scale and shift.
The prior-assisted construction (\prior) uses $r$ in two ways.

\textbf{Starting map.} We first align $r$ to the measurements.
After standardising $r$ over the image, $\tilde r=(r-\mu_r)/\sigma_r$, we fit a scale and shift by least squares in inverse depth,
\begin{equation}
    (a^\star,b^\star)=\arg\min_{a\ge 0,\,b}\;\sum_{i=1}^{N}\Big(a\,\tilde r(p_i)+b-\tfrac{1}{d_i}\Big)^2 ,
\end{equation}
and invert the result to obtain the aligned prior, a metric depth map $B(p)=1/\big(a^\star\tilde r(p)+b^\star\big)$.
To avoid dividing by values near zero, we clamp the denominator from below at half the smallest measured inverse depth.

\textbf{Residual label.} The TFM does not predict depth directly; it predicts how much the aligned prior must be corrected at each pixel.
The label of a measured pixel is the log ratio $y_i=\log d_i-\log B(p_i)$, and the prediction is $\hat D(p)=B(p)\exp(\hat y(p))$.
Where the prior is already correct, the TFM only has to predict zero.

\textbf{Columns.} $r$ also provides the columns.
Each row has 16 columns derived from the monocular network: position and $r$ (3); Sobel derivatives of $\tilde r$ and their magnitude (3); $\tilde r$ smoothed with Gaussians of $\sigma=4$ and $12$ pixels (2); and the first 8 principal components of its last-layer patch tokens (8).
We select this column set once, on development scenes of NYUv2 (\cref{sec:setup}), and use it on every dataset.

\subsection{Reading Out a Prediction}
\label{sec:prediction}

\paragraph{Median.}
Sparse depth from real sensors contains outliers: LiDAR returns from background surfaces that project onto foreground objects, mismatched structure-from-motion tracks, and flying pixels at depth edges.
A wrong measurement in the context widens the predictive distribution of nearby queries and pulls its mean towards the wrong value.
We therefore predict the median of $q(y\mid z,\mathcal{C})$, which is as accurate as the mean on clean measurements and far more robust when some are wrong (\cref{sec:robust}, Appendix).

\paragraph{Resolution.}
To reduce the number of query rows by a factor of four, we average the columns over $2\times2$ pixel blocks, predict one label per block, and upsample the predicted labels bilinearly to full resolution.

\paragraph{Context size.}
On KITTI frames with about 19K measurements, Kumo-Tabular is more accurate on random subsets of 2000 rows (1.31\,m on the validation split) than on the full context (2.19\,m), while TabICLv2 is designed for long contexts.
We use Kumo-Tabular~\cite{kumotabular} for $N\le 10{,}000$; when $N>2000$ we draw $\min(4,\lceil N/2000\rceil)$ random subsets of 2000 measurements, chosen to overlap as little as possible, run the TFM once per subset, and average the predictions.
For $N>10{,}000$ we use TabICLv2~\cite{tabiclv2}, which is built for long contexts, with all measurements in one context.
The rule depends on $N$ alone (\cref{sec:setup}).

\subsection{What Should We Expect?}
\label{sec:expect}

\cref{eq:pfn} says how the network is trained, but not how its error behaves as the context changes.
Nagler~\cite{nagler2023pfn} shows, for independent rows, that the variance of such a network vanishes as the context grows, and that its bias vanishes only when the prediction is sufficiently local to the query.
Pixels of one image are not independent, so we use the result only to derive four hypotheses that the experiments test.
\begin{itemize}
    \item[\textbf{H1}] \textbf{Proximity.} The error at a pixel increases with its distance from the nearest measurement. Distance in the image is our proxy for distance between rows.
    \item[\textbf{H2}] \textbf{Prior.} In the prior-assisted construction, a pixel far from every measurement is predicted about as well as the aligned prior predicts it; in the direct construction its error keeps growing.
    \item[\textbf{H3}] \textbf{Locality.} At a fixed context size, measurements near the query reduce the error more than measurements drawn at random.
    \item[\textbf{H4}] \textbf{Label agreement.} Measurements that disagree with the target depth cause errors that more context does not remove.
\end{itemize}

%% file: sec/4_experiments.tex
\section{Experiments}
\label{sec:experiments}

\subsection{Setup}
\label{sec:setup}

\paragraph{Benchmark.}
We use the zero-shot benchmark of OMNI-DC~\cite{zuo2025omnidc} and its released evaluation data, which fix the images, the input points, and the ground truth; baseline numbers are the published ones.
The sets with standard input points are NYUv2~\cite{silberman2012nyu} with 5 to 500 points (654 images), VOID~\cite{wong2020void} with 150 to 1500 points (800 images each), KITTI depth completion~\cite{uhrig2017kitti} with 8 to 64 LiDAR lines (749 images), and ETH3D~\cite{schops2017eth3d} with COLMAP points (390 images).
The synthetic-pattern sets apply ten sampling patterns to iBims-1~\cite{koch2018ibims}, ARKitScenes~\cite{baruch2021arkitscenes}, ETH3D, and DIODE~\cite{vasiljevic2019diode}: three random densities, two noise levels, ORB and SIFT keypoints, and three LiDAR scan patterns, averaged over the four indoor and the two outdoor subsets as in~\cite{zuo2025omnidc}.

\paragraph{Protocol.}
We report RMSE in metres over pixels with valid ground truth, per image and averaged over images~\cite{zuo2025omnidc}; MAE and relative error are in the Appendix.
We choose the prior-assisted columns, the median output, and the half-resolution prediction on 97 images from development scenes of NYUv2, and the context-size rule on the validation splits of KITTI and ETH3D.
These 97 images are NYUv2 test images (the Appendix gives results without them); no other test image is used for any choice.
To separate the tabular model from its inputs, we run nearest-neighbour regression, gradient-boosted trees, and a Gaussian process in the same pipeline, each tuned on the development scenes.

\subsection{Tabular Model versus Classical Regressors}
\label{sec:direct}

\begin{table}[t]
\centering
\caption{\textbf{The same table given to different regressors.} RMSE (m) on NYUv2, 654 test images, for 500, 200, and 100 measured points. Each classical regressor is tuned on development scenes. The first two blocks use no depth-trained network.}
\label{tab:regressors}
\small
\setlength{\tabcolsep}{3.2pt}
\begin{tabular}{llccc}
\toprule
Columns & Regressor & 500 & 200 & 100 \\
\midrule
\multirow{4}{*}{\shortstack[l]{position,\\colour}} & $k$-NN & .228 & .306 & .383 \\
 & boosted trees & .188 & .260 & .333 \\
 & Gaussian process & .185 & .264 & .336 \\
 & TFM & \best{.124} & \best{.183} & \best{.251} \\
\midrule
\multirow{4}{*}{\shortstack[l]{position, colour,\\DINOv2 features\\(\direct)}} & $k$-NN & .215 & .295 & .371 \\
 & boosted trees & .194 & .268 & .340 \\
 & Gaussian process & .168 & .233 & .302 \\
 & TFM & \best{.117} & \best{.167} & \best{.223} \\
\midrule
\multirow{4}{*}{\shortstack[l]{depth network\\(\prior)}} & $k$-NN & .152 & .182 & .205 \\
 & boosted trees & .143 & .177 & .205 \\
 & Gaussian process & .143 & .177 & .203 \\
 & TFM & \best{.108} & \best{.142} & \best{.169} \\
\bottomrule
\end{tabular}

\end{table}

\textbf{Without any depth-trained component, \direct{} reaches 0.117\,m on NYUv2 with 500 measurements.}
\cref{tab:regressors} gives the direct construction in its middle block.
For reference, OMNI-DC reaches 0.111\,m and G2-MonoDepth 0.122\,m on the same images and points (\cref{tab:nyu}).
With position and colour as the only columns the error is 0.124\,m: DINOv2 accounts for a small part of the result.

\textbf{Given identical inputs, tuned classical regressors have 32\% to 49\% higher error than the TFM.}
The best classical regressor is behind the TFM by 49\% with position and colour, by 44\% with DINOv2 features added, and by 32\% with the depth-network columns (\cref{tab:regressors}).
The gap is widest where the columns say least about depth.
With the \prior{} columns at 500 points, Kumo-Tabular (0.108\,m) is ahead of TabICLv2 (0.115\,m) and TabPFN~v2 (0.117\,m).

\subsection{Comparison with Zero-Shot Depth Completion}
\label{sec:benchmark}

\begin{table}[t]
\centering
\caption{\textbf{NYUv2.} RMSE (m), 654 test images, OMNI-DC evaluation data. Baselines are published numbers except the aligned DA-V2 row, which is the starting map of \prior{} in our run (published: .309 at 500 points). On the 557 test images not used for any choice, the error of \prior{} is lower than on all 654 at every density (.107 at 500 points; Appendix). $^\dagger$Published under its own protocol (500 random points, its own resolution), shown for reference.}
\label{tab:nyu}
\small
\setlength{\tabcolsep}{3.6pt}
{%
\begin{tabular}{lccccc}
\toprule
 & \multicolumn{5}{c}{NYUv2, number of points} \\
Method & 500 & 200 & 100 & 50 & 5 \\
\midrule
Marigold~\cite{ke2024marigold}, aligned & .426 & .428 & .431 & .436 & .545 \\
DA-V2~\cite{yang2024dav2}, aligned & .306 & .309 & .311 & .316 & .375 \\
G2-MonoDepth~\cite{wang2024g2} & .122 & .169 & .222 & .286 & .744 \\
OMNI-DC~\cite{zuo2025omnidc} & .111 & .147 & .180 & .225 & .536 \\
Marigold-DC$^\dagger$~\cite{viola2025marigolddc} & .124 & -- & -- & -- & -- \\
\midrule
\direct{} (no depth-trained network) & .117 & .167 & .223 & .308 & .956 \\
\prior{} & \best{.108} & \best{.142} & \best{.169} & \best{.202} & \best{.357} \\
\bottomrule
\end{tabular}}

\end{table}

\begin{table}[t]
\centering
\caption{\textbf{VOID, ETH3D with COLMAP points, and KITTI with real LiDAR.} RMSE (m); baselines are published numbers. $^\dagger$Marigold-DC~\cite{viola2025marigolddc} with its 10-member ensemble, evaluated in its own paper on the same VOID input maps. \direct{} reaches .541 on VOID 1500. The last row drops LiDAR returns that are more than 10\% deeper than another return within a $5\times7$ window before building the table (\cref{sec:failures}); the window was chosen on the KITTI validation split.}
\label{tab:real}
\small
\setlength{\tabcolsep}{3pt}
{%
\begin{tabular}{lccc@{\hskip 7pt}cc@{\hskip 7pt}ccc}
\toprule
 & \multicolumn{3}{c}{VOID} & \multicolumn{2}{c}{ETH3D SfM} & \multicolumn{3}{c}{KITTI (lines)} \\
Method & 1500 & 500 & 150 & in & out & 64 & 16 & 8 \\
\midrule
Marigold, aligned & .630 & .607 & .673 & .627 & 1.88 & 3.46 & -- & 3.50 \\
DA-V2, aligned & .605 & .582 & .644 & \best{.592} & 2.66 & 4.56 & -- & 4.69 \\
G2-MonoDepth & .568 & .574 & .691 & 1.07 & 2.45 & 1.61 & -- & 2.77 \\
OMNI-DC & .555 & .551 & .650 & .605 & \best{1.07} & \best{1.19} & \best{1.68} & \best{2.06} \\
Marigold-DC$^\dagger$ & .551 & .535 & .622 & -- & -- & -- & -- & -- \\
\midrule
\prior{} & \best{.525} & \best{.512} & \best{.587} & .664 & 1.81 & 1.50 & 2.36 & 3.14 \\
\quad with occlusion pre-filter & -- & -- & -- & -- & -- & 1.32 & 2.19 & 2.99 \\
\bottomrule
\end{tabular}}

\end{table}

\begin{table*}[t]
\centering
\caption{\textbf{Synthetic sampling patterns on four datasets whose test images were not used for any choice.} RMSE (m), averaged over subsets as in~\cite{zuo2025omnidc}. N5, N10: 5\% and 10\% noise; L64 to L8: LiDAR scan patterns. Baselines are published numbers. \direct{} reaches .093 indoors and 1.21 outdoors at 0.7\% density.}
\label{tab:synthetic}
\footnotesize
\setlength{\tabcolsep}{3.4pt}
\resizebox{\linewidth}{!}{%
\begin{tabular}{lcccccccccc@{\hskip 10pt}cccccccccc}
\toprule
 & \multicolumn{10}{c}{Indoor: ARKitScenes, iBims-1, ETH3D, DIODE} & \multicolumn{10}{c}{Outdoor: ETH3D, DIODE} \\
Method & 0.7\% & 0.1\% & 0.03\% & N5 & N10 & ORB & SIFT & L64 & L16 & L8 & 0.7\% & 0.1\% & 0.03\% & N5 & N10 & ORB & SIFT & L64 & L16 & L8 \\
\midrule
G2-MonoDepth & .107 & .195 & .327 & .134 & .155 & .438 & .402 & .882 & .150 & .219 & 1.45 & 2.26 & 3.24 & 1.55 & 1.66 & 2.02 & 1.84 & 1.20 & 1.76 & 2.40 \\
OMNI-DC & .084 & .151 & .233 & .090 & .097 & .240 & .203 & .611 & .107 & .163 & 1.28 & 1.89 & 2.48 & 1.32 & 1.39 & 1.65 & 1.43 & 1.27 & 1.51 & 2.10 \\
\midrule
$k$-NN, our columns & .114 & .165 & .203 & .271 & .382 & .189 & .182 & .100 & .131 & .154 & 1.69 & 2.36 & 2.78 & 2.46 & 3.09 & 1.84 & 1.96 & 1.55 & 1.95 & 2.41 \\
\prior{} & \best{.076} & \best{.108} & \best{.142} & \best{.078} & \best{.080} & \best{.139} & \best{.120} & \best{.073} & \best{.089} & \best{.110} & \best{1.19} & \best{1.78} & \best{2.23} & \best{1.22} & \best{1.23} & \best{1.37} & \best{1.37} & \best{1.08} & \best{1.42} & \best{1.82} \\
\bottomrule
\end{tabular}}

\end{table*}

\begin{figure*}[t]
\centering
\includegraphics[width=\linewidth]{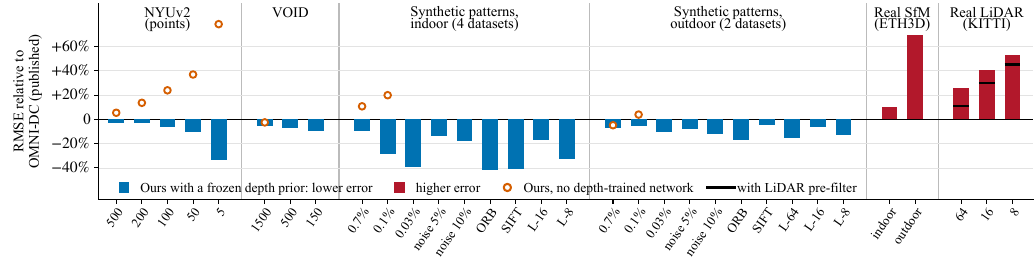}
\caption{\textbf{All benchmark settings at once.} Bars give the RMSE of \prior{} relative to the published OMNI-DC on the same data; a bar below zero is a lower error. Circles give \direct{} where it was run. Black lines give KITTI with the occlusion pre-filter. The method has lower error on NYUv2, VOID, and every synthetic-pattern setting shown, and higher error on real structure-from-motion points and real LiDAR, the two settings in which a few percent of the measurements are grossly wrong (\cref{sec:failures}).}
\label{fig:glance}
\end{figure*}

\textbf{\prior{} outperforms OMNI-DC on NYUv2 at every density.}
The margin is 3\% at 500 points and 33\% at 5 points (\cref{tab:nyu}).
The monocular network alone, aligned to the same 500 points, is at 0.306\,m, so the tabular model removes about two thirds of its error.
The direct construction is within 6\% of OMNI-DC at 500 points and falls behind as points become fewer: 5 labelled rows leave little to fit.

\textbf{The advantage holds on VOID and on synthetic patterns over four unseen datasets.}
\prior{} has lower error than OMNI-DC and the published Marigold-DC at all three VOID densities (\cref{tab:real}) and in all 20 synthetic-pattern settings, each an indoor or outdoor average (\cref{tab:synthetic}); \cref{fig:glance} shows every setting in one plot.
Indoors the margin grows as points become sparser, from 10\% at 0.7\% density to 39\% at 0.03\%, and is 42\% with ORB and 41\% with SIFT keypoints; outdoors it is 4\% with SIFT keypoints and 6\% to 15\% on the LiDAR scan patterns.

\textbf{OMNI-DC remains more accurate on real LiDAR and structure-from-motion points.}
On KITTI the error is 26\% to 53\% higher than OMNI-DC, and on ETH3D with COLMAP points 10\% to 70\% higher (\cref{tab:real}).
Both use measurements from a real pipeline, while the sets above sample them from the evaluation depth; \cref{sec:failures} examines why.

\subsection{Analysis: Distance, Prior, and Context}
\label{sec:analysis}

\begin{figure*}[t]
\centering
\includegraphics[width=\linewidth]{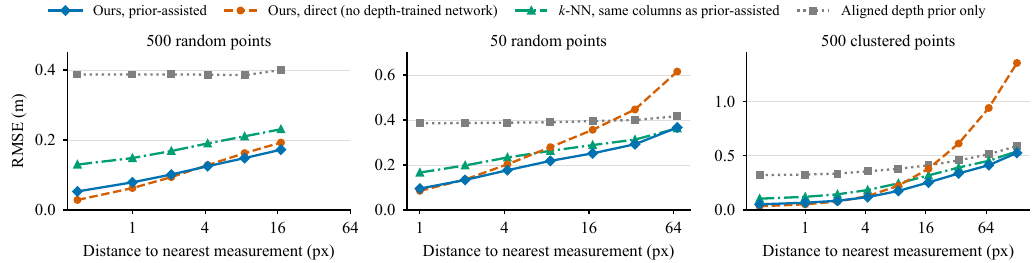}
\caption{\textbf{Error against distance to the nearest measurement.} NYUv2, 654 images, three point layouts. Near a measurement the direct construction is as accurate as the prior-assisted one. Far from every measurement the direct error keeps growing, while the prior-assisted error approaches that of the aligned monocular prior (dotted).}
\label{fig:distance}
\end{figure*}

\textbf{Error increases with distance from the nearest measurement (H1).}
\cref{fig:distance} bins every NYUv2 test pixel by its distance to the nearest measurement.
With 500 random points the prior-assisted error rises from 0.08\,m within 1.5 pixels to 0.17\,m beyond 12 pixels, and the same trend holds for 50 points and for clustered points.

\textbf{Far from the measurements, the depth network keeps the error bounded (H2).}
Within 3 pixels of a measurement the direct construction is as accurate as the prior-assisted one or more accurate (0.063\,m against 0.079\,m within 1.5 pixels, 500 random points).
With clustered points and beyond 96 pixels, the direct error is 1.35\,m, the prior-assisted error is 0.53\,m, and the aligned prior alone is at 0.59\,m.
This matches \cref{tab:nyu}: with 5 points most pixels are far from any measurement, so \prior{} (0.357\,m) is close to the aligned prior (0.375\,m) and \direct{} is at 0.956\,m.

\textbf{Nearby measurements are worth more than random ones (H3).}
On the NYUv2 development images with 500 points, the 250 measurements nearest to each query region give 0.115\,m, the same as all 500, and 250 drawn at random give 0.139\,m; on the KITTI validation split, 2000 nearest give 1.34\,m and 2000 random 1.67\,m.
On KITTI, all measurements in one TabICLv2 context are still more accurate (1.28\,m), so we use the full context.

\subsection{Robustness and Uncertainty}
\label{sec:robust}

\begin{figure*}[t]
\centering
\includegraphics[width=\linewidth]{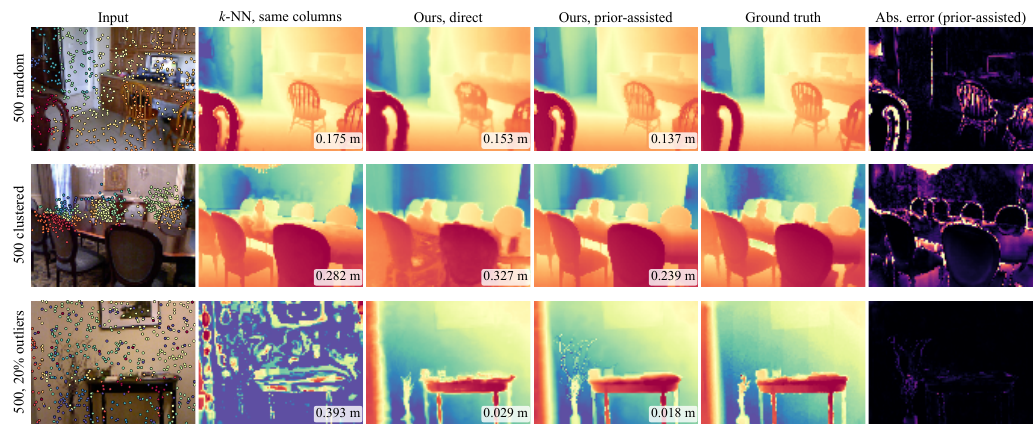}
\caption{\textbf{Three measurement layouts.} From top: 500 random points, 500 clustered points, and 500 points of which 20\% are replaced by random depths. Numbers are RMSE on the image. With clustered points the direct construction degrades away from the clusters. With corrupted points $k$-NN reproduces the outliers and the tabular model's median does not.}
\label{fig:qualitative}
\end{figure*}

\textbf{The median prediction is robust to randomly placed outliers.}
When 20\% of 500 points are replaced by random depths, the per-pixel error of \prior{} stays between 0.13\,m and 0.20\,m at every distance from the measurements, while $k$-NN on the same columns is at 0.5 to 0.6\,m (\cref{fig:qualitative}, bottom row).
On the benchmark's noise patterns the margin over OMNI-DC is 13\% at 5\% noise and 18\% at 10\% noise indoors (\cref{tab:synthetic}), larger than the 10\% margin on the same points without noise.

\textbf{The 90\% predictive interval is well calibrated on clean measurements.}
On NYUv2 pixels without a measurement, the interval between the 5\% and 95\% quantiles of \prior{} covers 89.5\% with 500 random points, 88.6\% with 50, and 88.4\% with 100 clustered points, and its mean width in log depth grows from 0.06 to 0.13 to 0.27.
With 20\% outliers the interval covers 99\%, so the interval is too wide under this corruption.

\subsection{Other Measurement Layouts}
\label{sec:layouts}

\begin{table}[t]
\centering
\caption{\textbf{Other layouts of the measurements, same method.} RMSE (m), NYUv2, 654 test images; we run the two baselines on identical inputs. Super-resolution: depth known at the centre of every $8\times8$ or $16\times16$ block. Missing square: depth known everywhere except a $76\times76$ square. Prior Depth Anything is unstable on the dense input with a hole under its public interface, and we leave the cell empty.}
\label{tab:layouts}
\small
\setlength{\tabcolsep}{4pt}
{%
\begin{tabular}{lccc}
\toprule
 & \multicolumn{2}{c}{Super-resolution} & Missing square \\
Method & $\times 8$ & $\times 16$ & (error inside) \\
\midrule
DA-V2, aligned & .298 & .297 & .273 \\
Prior Depth Anything~\cite{wang2025priorda} & .096 & .131 & -- \\
OMNI-DC~\cite{zuo2025omnidc} & .096 & .130 & .212 \\
\midrule
\direct{} (no depth-trained network) & \best{.083} & .142 & .176 \\
\prior{} & .086 & \best{.127} & \best{.128} \\
\bottomrule
\end{tabular}}

\end{table}

\textbf{The same method handles depth super-resolution and missing regions without modification.}
A low-resolution depth map and a depth map with a hole are tables with a different set of labelled rows (\cref{tab:layouts}).
At $\times 8$ super-resolution the direct construction has the lowest error; at $\times 16$ and inside the missing square the prior-assisted construction does, and the direct construction degrades for the reason of \cref{sec:analysis}.
As H2 predicts, the advantage of \prior{} over \direct{} grows with the size of the hole, from 11\% for a square of side 38 pixels to 43\% for a side of 152 pixels (Appendix).

\subsection{Failure Analysis on Real Measurements}
\label{sec:failures}

\textbf{The method loses only on sets whose measurements disagree with the reference depth (H4).}
On every set where \prior{} has lower error than OMNI-DC, the input points equal the reference depth or differ from it by benchmark noise.
On ETH3D with COLMAP points and on KITTI, 1.5\% to 3.6\% of the points that can be checked are off by more than 10\%, which alone gives the measurements an RMSE of 1.0 to 3.1\,m (Appendix).
Layout does not separate the groups: LiDAR scan patterns appear in both.

\begin{table}[t]
\centering
\caption{\textbf{Replacing values at fixed checked pixels reduces RMSE by 75\% to 80\% on ETH3D and by 21\% on KITTI validation.} RMSE (m) of \prior{}. \emph{Checked pixels}: only the released points that fall on a pixel with reference depth. \emph{Same pixels}: those same points with their values replaced by the reference depth; this column and the previous one use identical locations. \emph{Random pixels}: as many points as released, at random pixels with reference depth. KITTI: 160 validation frames. The random-pixel column uses 132 and 131 of the ETH3D images and all 251 KITTI validation frames.}
\label{tab:control}
\small
\setlength{\tabcolsep}{3.2pt}
{%
\begin{tabular}{lcccc}
\toprule
 & Released & Released values, & Reference values, & Reference values, \\
Input points & points & checked pixels & same pixels & random pixels \\
\midrule
ETH3D, COLMAP, indoor & .664 & .660 & .136 & .099 \\
ETH3D, COLMAP, outdoor & 1.81 & 2.46 & .617 & .375 \\
KITTI val, 64 lines & 1.29 & .886 & .701 & .589 \\
\bottomrule
\end{tabular}}

\end{table}

\textbf{At the same pixels, reference values reduce the error by 75\% to 80\% on ETH3D and by 21\% on KITTI.}
We restrict the released points to pixels that have reference depth, and then replace their values by the reference depth, leaving every location unchanged (\cref{tab:control}, second and third columns).
On ETH3D the result (.136 indoor, .617 outdoor) is below published OMNI-DC with the released points (.605 and 1.07).
Random pixels with reference values (fourth column) change location, count and, on some sets, the images, so we do not compare that column with the third.
On KITTI the restriction to checked pixels is itself informative: the reference depth keeps a LiDAR return only if it agrees with stereo and with neighbouring frames, and of the removed returns that lie next to a reference pixel, 17\% differ from it by more than 10\%, against 1.6\% of the kept ones.
Removing them lowers the error from 1.29 to .886 with no value changed.
Because the control uses reference depth, we report these numbers as a diagnosis only: they implicate disagreement between measured values and the target (H4) as the main limitation on these sets.

\begin{table}[t]
\centering
\caption{\textbf{Clustered corruption raises the error of the tabular model far more than scattered corruption.} RMSE (m) on NYUv2 development images with 500 measurements, of which 6\% or 20\% are made 1.5 times deeper, either at random points or in spatial clusters of about ten neighbouring points. 97 development images.}
\label{tab:coherent}
\small
\setlength{\tabcolsep}{4pt}
{%
\begin{tabular}{lccccc}
\toprule
 & & \multicolumn{2}{c}{6\% corrupted} & \multicolumn{2}{c}{20\% corrupted} \\
Method & Clean & scattered & clustered & scattered & clustered \\
\midrule
Aligned DA-V2 & .318 & .322 & .324 & .382 & \best{.396} \\
$k$-NN, our columns & .168 & .224 & \best{.286} & .368 & .487 \\
\direct{} & .127 & .135 & .344 & .148 & .610 \\
\prior{} & \best{.115} & \best{.122} & .338 & \best{.129} & .605 \\
\bottomrule
\end{tabular}}

\end{table}

\textbf{Scattered errors are tolerated; clustered errors are not.}
\cref{sec:robust} showed that the median prediction survives 20\% random outliers, yet a few percent of wrong returns on KITTI are enough to lose.
An intervention on NYUv2 tests whether spatial clustering can produce this sensitivity.
We corrupt the same number of NYUv2 measurements by the same factor in two ways: at points chosen at random, and in clusters of about ten neighbouring points (\cref{tab:coherent}).
With 6\% of the points corrupted at random the error rises from .115 to .122.
With the same 6\% in clusters it rises to .338, worse than nearest-neighbour regression on the same table and than the aligned prior alone.
A wrong label surrounded by correct ones is an outlier under the predictive distribution, and the median discounts it; a cluster of wrong labels that agree with each other is, to a regressor, a surface.
See-through LiDAR returns and mismatched structure-from-motion tracks can produce errors of the second kind.

\textbf{A simple occlusion filter on KITTI recovers part of the gap.}
On KITTI the clustered errors have a known geometric cause: the LiDAR sits above the camera, so some returns from the background project onto foreground objects.
A standard test that uses no reference depth drops a return that is more than 10\% deeper than another return within a $5\times7$ pixel window.
With the window chosen on the validation split, the test error falls by 12\% at 64 lines, 7\% at 16, and 5\% at 8 (\cref{tab:real}).
The test finds about 60\% of the disagreeing returns, and the results remain behind OMNI-DC.
The test is specific to projected LiDAR; every other benchmark number uses the measurements as released.

Changes that did not reduce the error on development data, among them a metric depth network as the starting map, local alignment, an additive target, and averaging two tabular models, are listed in the Appendix.

%% file: sec/6_conclusion.tex
\section{Limitations and Conclusions}
\label{sec:conclusion}

\paragraph{Limitations.}
TabDepth trusts its measurements: on raw KITTI LiDAR and structure-from-motion points, where a few percent of the measured values disagree with the true depth, OMNI-DC remains more accurate (\cref{tab:real,tab:control}).
The method also needs measurements in the image it predicts; a first experiment with labelled rows taken from other images is in the Appendix.

\paragraph{Conclusion.}
A tabular foundation model trained only on synthetic tables, given the pixels of an image as rows and a few hundred measured depths as labels, predicts dense depth that is competitive with networks trained for depth completion, as long as the measurements cover the scene and agree with it. This result suggests that in-context regression is a viable alternative to task-specific training for dense prediction from sparse measurements.

%% file: sec/X_suppl.tex
\appendix
\crefalias{section}{appendix}
\newcommand{\BANKN}{28}
\renewcommand{\thetable}{S\arabic{table}}
\renewcommand{\thefigure}{S\arabic{figure}}
\setcounter{table}{0}\setcounter{figure}{0}

\noindent This appendix contains the material that the main text refers to as the Appendix: the other error measures (\cref{sec:supp_metrics}), the comparison of the released input points with ground truth (\cref{sec:supp_audit}), details of the diagnostics (\cref{sec:supp_diag}), the selection of the LiDAR pre-filter (\cref{sec:supp_occ}), the experiment with context rows from other images (\cref{sec:supp_bank}), changes that did not reduce the error (\cref{sec:supp_negative}), and more qualitative results (\cref{sec:supp_qual}).
Tables and figures of the appendix are numbered S1, S2, and so on; numbers without a letter refer to the main text.
\prior{} is the prior-assisted construction and \direct{} the direct construction, as in the main text.

\section{MAE and Relative Error}
\label{sec:supp_metrics}

The main text reports RMSE.
The OMNI-DC paper~\cite{zuo2025omnidc} also publishes one second measure per set: relative error (REL) on NYUv2 and ETH3D, MAE on VOID and KITTI, and both on the synthetic patterns.
\cref{tab:supp_real,tab:supp_synth} compare on those measures.
REL is the mean of $|\hat D-D|/D$ over pixels with ground truth $D$.
The ranking is the same as under RMSE.
\prior{} has lower MAE and lower REL than OMNI-DC in all 20 synthetic-pattern settings, lower REL at every point count on NYUv2, and lower MAE at every density on VOID.
It has higher error than OMNI-DC on KITTI and on ETH3D with COLMAP points, the two sets on which it also loses under RMSE.

\begin{table}[h]
\centering
\caption{\textbf{Real-measurement sets under the second published measure.} Baseline numbers are the published ones~\cite{zuo2025omnidc}. Image counts as in Tabs.~2 and 3 of the main text.}
\label{tab:supp_real}
\small
\setlength{\tabcolsep}{3.6pt}
{%
\begin{tabular}{lccccc}
\toprule
NYUv2, REL & 500 & 200 & 100 & 50 & 5 \\
\midrule
G2-MonoDepth & .017 & .027 & .038 & .056 & .207 \\
OMNI-DC & .014 & .021 & .029 & .041 & .142 \\
\direct{} & .015 & .025 & .038 & .061 & .286 \\
\prior{} & \best{.013} & \best{.019} & \best{.025} & \best{.031} & \best{.076} \\
\bottomrule
\end{tabular}}

\vspace{4pt}
{%
\begin{tabular}{lccc@{\hskip 10pt}ccc}
\toprule
 & \multicolumn{3}{c}{VOID, MAE (m)} & \multicolumn{3}{c}{KITTI, MAE (m)} \\
Method & 1500 & 500 & 150 & 64 lines & 16 lines & 8 lines \\
\midrule
G2-MonoDepth & .159 & .182 & .247 & .376 & -- & .901 \\
OMNI-DC & .150 & .164 & .211 & \best{.270} & \best{.441} & \best{.597} \\
\prior{} & \best{.137} & \best{.143} & \best{.172} & .355 & .777 & 1.16 \\
\quad with occlusion pre-filter & -- & -- & -- & .298 & .670 & 1.05 \\
\bottomrule
\end{tabular}}

\vspace{4pt}
\begin{tabular}{lcc}
\toprule
ETH3D, COLMAP points, REL & Indoor & Outdoor \\
\midrule
OMNI-DC & \best{.090} & \best{.053} \\
\prior{} & .098 & .064 \\
\bottomrule
\end{tabular}

\end{table}

\begin{table*}[t]
\centering
\caption{\textbf{Synthetic patterns under MAE and relative error.} Same settings and layout as Tab.~4 of the main text; OMNI-DC numbers are the published ones. Before rounding, \prior{} is lower in every column, including the indoor N5 column where both round to .004.}
\label{tab:supp_synth}
\footnotesize
\setlength{\tabcolsep}{3.4pt}
\resizebox{\linewidth}{!}{%
\begin{tabular}{llcccccccccc@{\hskip 10pt}cccccccccc}
\toprule
 & & \multicolumn{10}{c}{Indoor: ARKitScenes, iBims-1, ETH3D, DIODE} & \multicolumn{10}{c}{Outdoor: ETH3D, DIODE} \\
 & Method & 0.7\% & 0.1\% & 0.03\% & N5 & N10 & ORB & SIFT & L64 & L16 & L8 & 0.7\% & 0.1\% & 0.03\% & N5 & N10 & ORB & SIFT & L64 & L16 & L8 \\
\midrule
\multirow{2}{*}{MAE (m)} & OMNI-DC & .015 & .038 & .076 & .016 & .019 & .127 & .101 & .016 & .024 & .050 & .292 & .599 & .970 & .313 & .341 & .514 & .403 & .303 & .412 & .715 \\
 & \prior{} & \best{.013} & \best{.024} & \best{.037} & \best{.014} & \best{.015} & \best{.058} & \best{.045} & \best{.012} & \best{.018} & \best{.028} & \best{.256} & \best{.447} & \best{.668} & \best{.262} & \best{.268} & \best{.374} & \best{.317} & \best{.230} & \best{.331} & \best{.503} \\
\midrule
\multirow{2}{*}{REL} & OMNI-DC & .004 & .010 & .020 & .004 & .005 & .057 & .046 & .004 & .006 & .014 & .022 & .044 & .070 & .023 & .024 & .039 & .034 & .023 & .031 & .048 \\
 & \prior{} & \best{.003} & \best{.006} & \best{.009} & \best{.004} & \best{.004} & \best{.023} & \best{.018} & \best{.003} & \best{.005} & \best{.008} & \best{.019} & \best{.033} & \best{.049} & \best{.019} & \best{.019} & \best{.026} & \best{.025} & \best{.017} & \best{.024} & \best{.033} \\
\bottomrule
\end{tabular}}

\end{table*}

\paragraph{NYUv2 without the development images.}
The 97 development images on which we chose the columns of \prior{}, the median output and the prediction stride are part of the 654-image NYUv2 test set.
\cref{tab:supp_nodev} repeats the NYUv2 comparison on the 557 test images that were not used for any choice.
The errors are slightly lower than on all 654 images, for both constructions and at every point count, because the development images are harder than average (the aligned prior alone is at .331 on them and at .302 on the rest).
The selection therefore did not favour the numbers of the main text.
All other datasets are untouched by any choice.

\begin{table}[h]
\centering
\caption{\textbf{NYUv2 with and without the development images.} RMSE (m). The published baselines are available on all 654 images only (OMNI-DC: .111, .147, .180, .225, .536).}
\label{tab:supp_nodev}
\small
\setlength{\tabcolsep}{4pt}
\begin{tabular}{llccccc}
\toprule
Method & Images & 500 & 200 & 100 & 50 & 5 \\
\midrule
\multirow{2}{*}{\direct{}} & all 654 & .117 & .167 & .223 & .308 & .956 \\
 & 557 unused & .115 & .166 & .222 & .307 & .956 \\
\midrule
\multirow{2}{*}{\prior{}} & all 654 & .108 & .142 & .169 & .202 & .357 \\
 & 557 unused & .107 & .141 & .168 & .200 & .348 \\
\bottomrule
\end{tabular}

\end{table}

\section{The Released Input Points}
\label{sec:supp_audit}

\input{sec/X_audit_table}

\cref{tab:audit} compares the released input points with the ground truth at the same pixels; no model is involved.
A point counts as off when it differs from the ground truth by more than 10\%.
Only points that fall on a pixel with ground truth can be checked.
The sets above the rule are sampled from the ground truth, so their points are exact, except for the noise pattern, which is corrupted by construction.
The sets below the rule are measured independently of the ground truth, and 1.5\% to 3.6\% of their checkable points are off.
These are the sets on which \prior{} loses to OMNI-DC.
The last column shows that the spatial layout does not separate the two groups: KITTI with 64 lines is the densest set in the table, and the ORB pattern, on which \prior{} wins, is as sparse as the COLMAP points.

On KITTI the check above understates the problem, because the ground truth was itself built by keeping the LiDAR returns that are consistent over several frames.
Of the raw returns, 38\% fall on pixels without ground truth.
Among those that lie next to a ground-truth pixel, 16.8\% differ from it by more than 10\%, and 14.1\% are deeper, the signature of a return from the background seen through or beside a foreground object.
For the returns that fall on a ground-truth pixel the share is 1.6\%.

\input{sec/X_properties_table}

\cref{tab:properties} summarises what each family of methods requires and returns.

\section{Details of the Diagnostics}
\label{sec:supp_diag}

\paragraph{Error against distance.}
\cref{tab:supp_dist} gives the numbers behind Fig.~4 of the main text for four of the distance bins.

\begin{table}[h]
\centering
\caption{\textbf{RMSE (m) by distance to the nearest measurement,} NYUv2, 654 test images. Next to a measurement \direct{} is at least as accurate as \prior{}; far from every measurement its error keeps growing, while the error of \prior{} stays below that of the aligned prior (H1 and H2).}
\label{tab:supp_dist}
\small
\setlength{\tabcolsep}{3.4pt}
{%
\begin{tabular}{llcccc}
\toprule
 & & \multicolumn{4}{c}{Distance to the nearest measurement (px)} \\
Points & Method & 0.5--1.5 & 6--12 & 24--48 & 48--96 \\
\midrule
\multirow{4}{*}{50 random} & Aligned prior & .385 & .390 & .400 & .417 \\
 & $k$-NN, same columns & .166 & .263 & .313 & \best{.363} \\
 & \direct{} & \best{.085} & .279 & .447 & .615 \\
 & \prior{} & .095 & \best{.218} & \best{.292} & .367 \\
\midrule
\multirow{4}{*}{500 clustered} & Aligned prior & .325 & .380 & .457 & .513 \\
 & $k$-NN, same columns & .121 & .245 & .391 & .453 \\
 & \direct{} & \best{.052} & .223 & .612 & .939 \\
 & \prior{} & .067 & \best{.175} & \best{.336} & \best{.413} \\
\bottomrule
\end{tabular}}

\end{table}

\paragraph{Median and mean.}
The method reports the median of the predictive distribution.
On NYUv2 (654 images, 500 random points) the mean of the same distribution is as accurate as the median when the measurements are clean (.108 and .109), and much worse when 20\% of them are replaced by random depths (.291 against .130).

\paragraph{Size of the missing region.}
\cref{tab:supp_hole} gives the sweep over the size of the missing square that Sec.~4.6 of the main text summarises.

\begin{table}[h]
\centering
\caption{\textbf{RMSE (m) inside a missing square,} NYUv2; depth is known everywhere else. Images are $304\times228$, so the squares cover 2\%, 8\% and 33\% of the image. The middle size is the one of Tab.~5 of the main text (654 images); the other two sizes use 574 of the 654 images.}
\label{tab:supp_hole}
\small
\setlength{\tabcolsep}{5pt}
\begin{tabular}{lccc}
\toprule
 & \multicolumn{3}{c}{Side of the missing square (px)} \\
Method & 38 & 76 & 152 \\
\midrule
Aligned prior & .229 & -- & .319 \\
\direct{} & .083 & .176 & .330 \\
\prior{} & \best{.074} & \best{.128} & \best{.187} \\
\bottomrule
\end{tabular}

\end{table}

\paragraph{Image-feature columns at several scales.}
We tested whether image features pooled over larger windows help when measurements are few.
We add to each row the first 8 principal components of the DINOv2 tokens averaged over $3\times3$ and over $7\times7$ token windows (16 columns).
On the 97 development images the error of \direct{} falls from .328 to .312 with 50 random points and from .697 to .665 with 100 clustered points, and is unchanged with 500 random points (.127).
For \prior{} the change is at most .002 in all three settings.
The gain is confined to the sparse settings of the direct construction, so we keep the smaller column set.

\section{Selecting the LiDAR Pre-Filter}
\label{sec:supp_occ}

The pre-filter of Sec.~4.7 of the main text drops a LiDAR return when it is more than 10\% deeper than the shallowest return inside a window centred on it.
It uses the projected returns only: no image, no ground truth, and no model.
\cref{tab:supp_occ} shows the three window sizes that we compared on the KITTI validation split.
We chose the $5\times7$ window on the 64-line setting, where the differences are largest, and used it for all three settings on the test split.
On the validation split a $7\times11$ window is slightly better at 16 and 8 lines; we did not use a second setting.
In an offline check against the ground truth of the validation split, the chosen window removes about 60\% of the returns that are off by more than 10\%, and 68\% of the returns that it removes are of that kind.

\begin{table}[h]
\centering
\caption{\textbf{Window of the occlusion pre-filter, validation split.} RMSE (m) of \prior{}. The $5\times7$ window is the one used on the test split.}
\label{tab:supp_occ}
\small
\setlength{\tabcolsep}{5pt}
\begin{tabular}{lcccc}
\toprule
 & & \multicolumn{3}{c}{Window (rows $\times$ columns)} \\
KITTI validation & None & $5\times7$ & $7\times11$ & $9\times15$ \\
\midrule
64 lines (63 frames) & 1.29 & \best{1.12} & 1.26 & 1.59 \\
16 lines (23 frames) & 2.08 & 2.00 & \best{1.94} & 1.97 \\
8 lines (64 frames) & 2.86 & 2.73 & \best{2.69} & 2.72 \\
\bottomrule
\end{tabular}

\end{table}

\section{Context Rows from Other Images}
\label{sec:supp_bank}

In the main text every labelled row comes from the image that is being predicted.
Here the query image has no measurement at all, and the labelled rows come from images of other scenes.

\paragraph{Setup.}
We use the NYUv2 test images.
The bank holds one image from each of up to 64 scenes, with 300 labelled pixels drawn at random from each image; the queries are images of scenes that are not in the bank.
The label is the log depth.
A row cannot use a per-image alignment to measurements, so the columns must carry the scale of the scene.
We compare two column sets: position, colour, and 32 principal components of the DINOv2 tokens; and the same with 16 components of the tokens averaged over a $5\times5$ token window and 16 components of the image-mean token, which describe the neighbourhood and the scene.
Principal components are fitted on the bank images only.
Kumo-Tabular receives 2{,}000 rows drawn at random from the bank; the other regressors receive all rows.

\begin{table}[h]
\centering
\caption{\textbf{No measurement in the query image.} RMSE (m) on 50 to 52 NYUv2 images from scenes that are not in the bank. The last row is a network trained for metric depth on indoor data, shown for reference.}
\label{tab:supp_bank}
\small
\setlength{\tabcolsep}{4pt}
{%
\begin{tabular}{llccc}
\toprule
 & & \multicolumn{3}{c}{Images in the bank} \\
Columns & Regressor & 4 & 16 & 64 \\
\midrule
\multirow{4}{*}{Pixel features} & ridge regression & .958 & .913 & .816 \\
 & boosted trees & 1.14 & .949 & .802 \\
 & TabICLv2 & 1.31 & 1.24 & .944 \\
 & Kumo-Tabular & 1.48 & 1.23 & .863 \\
\midrule
\multirow{4}{*}{\shortstack[l]{+ neighbourhood\\and scene features}} & ridge regression & .750 & 1.44 & .620 \\
 & boosted trees & 1.17 & .876 & .712 \\
 & TabICLv2 & .891 & .988 & .721 \\
 & Kumo-Tabular & .975 & .901 & \best{.619} \\
\midrule
\multicolumn{2}{l}{Median depth of the bank (constant)} & 1.00 & 1.00 & 1.00 \\
\multicolumn{2}{l}{DA-V2, scale and shift fitted on the bank} & .696 & .739 & .746 \\
\multicolumn{2}{l}{DA-V2 metric indoor network} & \best{.652} & \best{.644} & .644 \\
\bottomrule
\end{tabular}}

\end{table}

\paragraph{Result.}
\cref{tab:supp_bank} shows three things.
First, from 4 to 64 images the error falls for every regressor and both column sets, although not at every step.
Second, the neighbourhood and scene columns matter more than the choice of regressor: with 64 images they reduce the error of Kumo-Tabular from .863 to .619.
Third, with 64 images Kumo-Tabular and ridge regression are tied (.619 and .620), so in this setting the tabular model adds nothing to the features, and both are close to a network trained for metric depth (.644).

\paragraph{A larger bank.}
A second run enlarges the bank to 128 scenes and up to two images per scene (244 images, 73{,}200 labelled rows), and lets Kumo-Tabular use more of the rows by averaging its predictions over 16 random subsets of 2{,}000 rows.
\cref{tab:supp_bank2} gives the result on \BANKN{} queries; the run is slow, and we stopped it at this number.
The error keeps falling with the size of the bank, and with 244 images Kumo-Tabular is below ridge regression on the same bank and below the metric network.
The number of queries is small, and we therefore state the conclusion of the main text on the 50-query run above.

\begin{table}[h]
\centering
\caption{\textbf{A larger bank,} neighbourhood and scene columns, RMSE (m) on \BANKN{} NYUv2 queries from scenes that are not in the bank. The queries differ from those of \cref{tab:supp_bank}.}
\label{tab:supp_bank2}
\small
\setlength{\tabcolsep}{5pt}
\begin{tabular}{lccc}
\toprule
 & \multicolumn{3}{c}{Images in the bank} \\
Regressor & 64 & 128 & 244 \\
\midrule
ridge regression & .680 & .626 & .587 \\
boosted trees & .672 & .604 & .567 \\
Kumo-Tabular, 1 subset & .667 & .617 & .548 \\
Kumo-Tabular, 16 subsets & \best{.642} & \best{.585} & \best{.508} \\
\midrule
DA-V2 metric indoor network & .756 & .756 & .756 \\
\bottomrule
\end{tabular}

\end{table}

\section{Changes That Did Not Reduce the Error}
\label{sec:supp_negative}

We list the changes that we tried on development data and did not adopt, so that others need not repeat them.
None of them reduced the error of \prior{} on the NYUv2 development images or the KITTI validation split.
\begin{itemize}
    \item \textbf{A metric depth network as the starting map,} aligned by scale only, in place of the relative-depth network aligned by scale and shift.
    \item \textbf{Local alignment} of the prior to nearby measurements before the tabular model, and a weighted global alignment. The weighted alignment improves the starting map on KITTI with 8 lines, but almost none of the gain remains after the tabular model.
    \item \textbf{An additive target} (depth minus aligned prior) in place of the log residual.
    \item \textbf{Columns built from the measurements,} such as the residuals of the nearest measurements.
    \item \textbf{Fewer columns when measurements are few.}
    \item \textbf{Averaging Kumo-Tabular and TabICLv2,} and using more ensemble members inside either model.
    \item \textbf{A context of nearby measurements only.} On NYUv2 with 500 points the 250 nearest measurements match the full context (.115) and 250 random ones do not (.139), which supports H3; but selecting neighbours does not improve on the full context, and on KITTI 2{,}000 nearest measurements per region (1.34) are worse than all measurements in one TabICLv2 context (1.28).
\end{itemize}

\section{More Qualitative Results}
\label{sec:supp_qual}

\begin{figure*}[t]
\centering
\includegraphics[width=.92\linewidth]{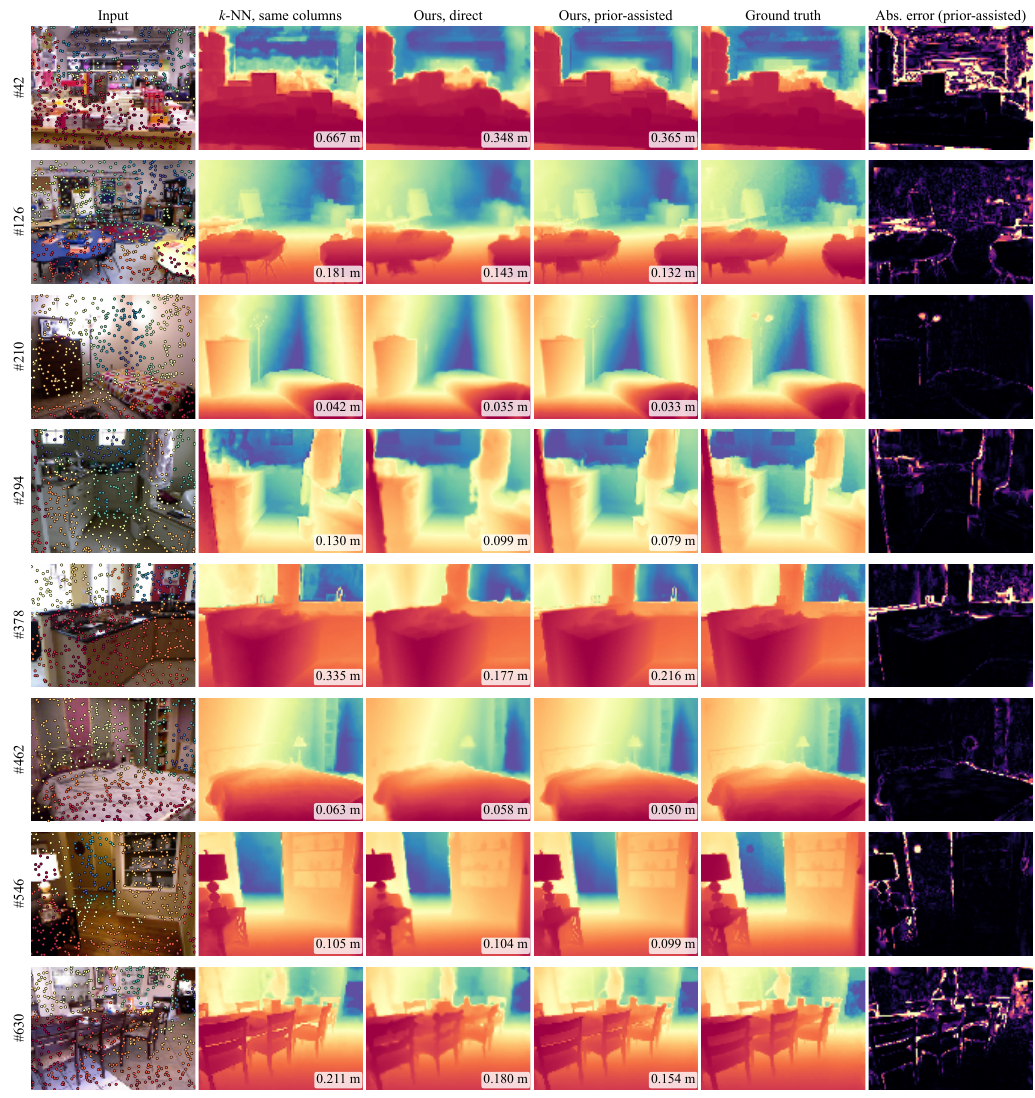}
\caption{\textbf{500 measurements,} eight NYUv2 test images at a fixed stride (image index on the left). Numbers are the RMSE of each map. Depth colours share one range per row; the error map is clipped at 0.5\,m.}
\label{fig:supp_q500}
\end{figure*}

\begin{figure*}[t]
\centering
\includegraphics[width=.92\linewidth]{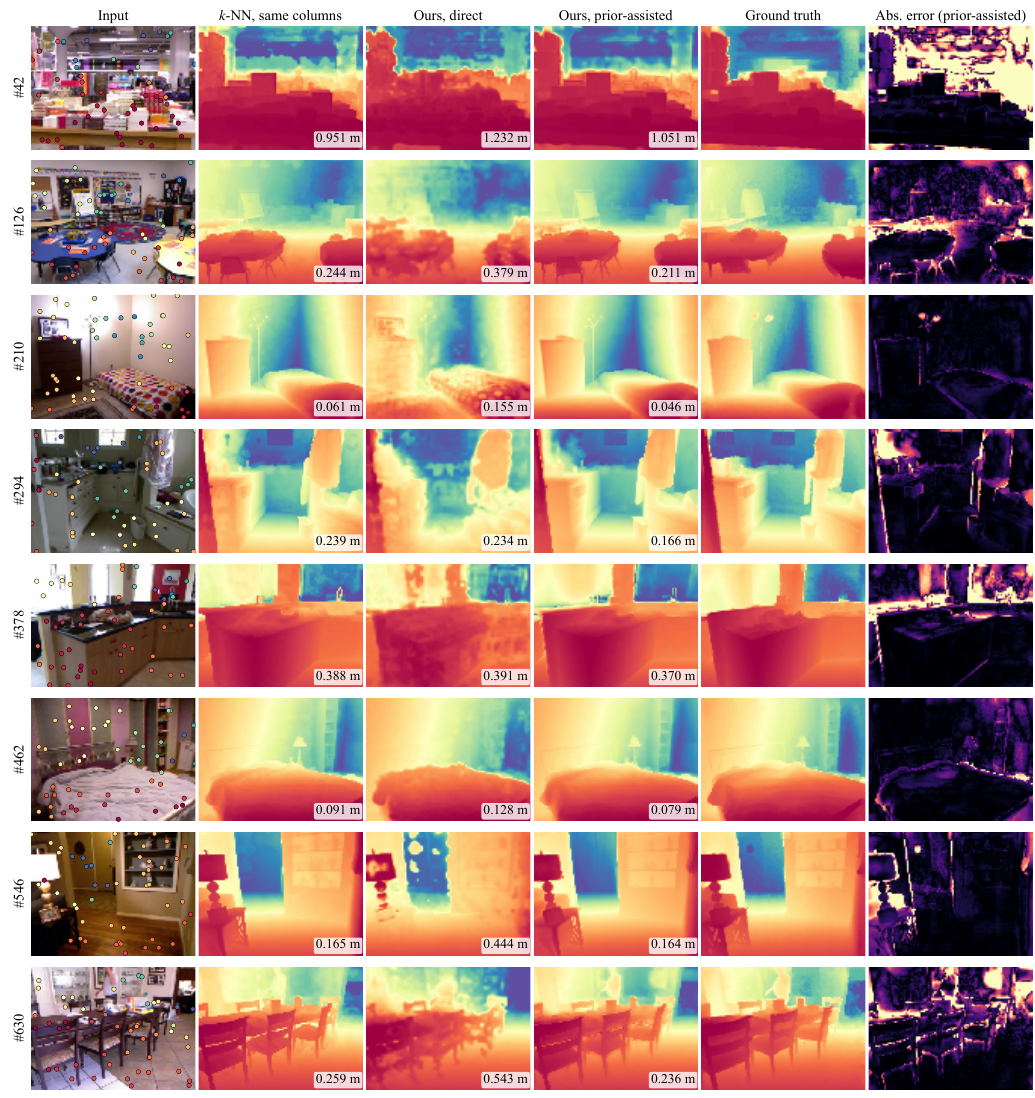}
\caption{\textbf{50 measurements,} same images and layout as \cref{fig:supp_q500}.}
\label{fig:supp_q50}
\end{figure*}

\cref{fig:supp_q500,fig:supp_q50} show eight NYUv2 test images with 500 and with 50 measurements.
The images are taken at a fixed stride through the test order and are not selected by result.
With 500 measurements the two constructions are hard to tell apart by eye.
With 50 the direct construction shows the failure that H2 predicts: regions far from every measurement take the depth of the nearest labelled rows with similar features, which can be wrong by a metre, while the prior-assisted construction keeps the shape of the monocular prior there.
With 50 measurements the first row is a failure of every method shown.

%% file: sec/X_audit_table.tex
\begin{table}[t]
\centering
\caption{\textbf{The released input points compared with ground truth, no model involved.} The sets on which the method loses (bottom) are those whose points contain a small share of gross errors. Their spatial layout is similar to sets on which the method wins.}
\label{tab:audit}
\small
\setlength{\tabcolsep}{3.6pt}
{%
\begin{tabular}{lccc}
\toprule
 & Points off & RMSE of the & Median distance \\
Input points & by $>$10\% & points (m) & to a point (px) \\
\midrule
NYUv2, 500 & 0\% & 0 & 5.6 \\
VOID, 1500 & 0\% & 0 & 11.0 \\
ETH3D outdoor, ORB pattern & 0\% & 0 & 20.2 \\
ETH3D outdoor, L64 pattern & 0\% & 0 & 3.6 \\
iBims-1, 10\% noise pattern & 7.7\% & 0.51 & 5.7 \\
\midrule
ETH3D, COLMAP, indoor & 3.6\% & 1.05 & 20.1 \\
ETH3D, COLMAP, outdoor & 2.9\% & 3.11 & 15.1 \\
KITTI, 64 lines & 1.5\% & 1.71 & 1.4 \\
KITTI, 8 lines & 1.6\% & 1.69 & 9.8 \\
\bottomrule
\end{tabular}}

\end{table}

%% file: sec/X_properties_table.tex
\begin{table}[t]
\centering
\caption{\textbf{What each family of methods requires and returns.}}
\label{tab:properties}
\small
\setlength{\tabcolsep}{7pt}
\begin{tabular}{lcccc}
\toprule
 & Completion & Test-time & Depth-trained & Predictive \\
Method & training & optimisation & network & distribution \\
\midrule
OMNI-DC, G2-MonoDepth, PriorDA & yes & no & yes & no \\
Marigold-DC, test-time adaptation & no & yes & yes & no \\
\prior{} (prior-assisted) & no & no & yes (frozen) & yes \\
\direct{} (direct) & no & no & no & yes \\
\bottomrule
\end{tabular}
\end{table}